# Enhancing Pothole Detection and Characterization: Integrated Segmentation and Depth Estimation in Road Anomaly Systems

Uthman Baroudi*, Alala BaHamid, Yasser Elalfy and Ziad Al Alami

Computer Engineering Department
Interdisciplinary Research Center for Intelligent Secure Systems
Dhahran, Saudi Arabia
* Corresponding Author

**Abstract**

Road anomaly detection plays a crucial role in road maintenance and in enhancing the safety of both drivers and vehicles. Recent machine learning approaches for road anomaly detection have overcome the tedious and time-consuming process of manual analysis and anomaly counting; however, they often fall short in providing a complete characterization of road potholes. In this paper, we leverage transfer learning by adopting a pre-trained YOLOv8-seg model for the automatic characterization of potholes using digital images captured from a dashboard-mounted camera. Our work includes the creation of a novel dataset, comprising both images and their corresponding depth maps, collected from diverse road environments in Al-Khobar city and the KFUPM campus in Saudi Arabia. Our approach performs pothole detection and segmentation to precisely localize potholes and calculate their area. Subsequently, the segmented image is merged with its depth map to extract detailed depth information about the potholes. This integration of segmentation and depth data offers a more comprehensive characterization compared to previous deep learning-based road anomaly detection systems. Overall, this method not only has the potential to significantly enhance autonomous vehicle navigation by improving the detection and characterization of road hazards but also assists road maintenance authorities in responding more effectively to road damage.

## 1  Introduction

Road surface anomaly detection and characterization have emerged as essential areas of research in the transportation systems, driven by the demand to enhance road safety, minimize damage to vehicles, and optimize maintenance operations. Road anomalies such as potholes and cracks pose a substantial hazard to drivers and vehicles and may result in life loss and vehicle damage. Road anomaly detection has undergone significant progress, driven by the advances in sensing technologies, deep learning, and computer vision. In comparison with traditional detection methods that involve manual inspection associated with cost and time consumption, deep learning has automated and revolutionized road anomaly detection and streams analysis of videos and images at unprecedented accuracies where large-scale datasets enable development of robust anomaly systems [1]. Most of available datasets captured road images in either horizontal view in which models need to prioritize the road region over irrelevant areas, or top-down view that categorize damage and background [2]. Murty et al. [3] proposed pothole detection model using Convolutional Neural Networks. ResNetV2, ResNet50, VGG19, and YOLOv8, were used for pothole detection in road images. Zhang et al. [4] improved the detection accuracy of cracks and potholes by integrating YOLO v3 model with a multi-level attention mechanism. Majidifrad et al. [1] combined YOLO and U-Net models to identify road anomalies and then categorize their severity, simultaneously. The crack density per pavement defect is determined by combining the results. However, the previous models fail to provide full characterisation of pothole anomaly which is crucial for road maintenance authorities and for alerting drivers on the severity level. Therefore, this paper developed an anomaly detection Yolov8 based system to provide comprehensive details on potholes. Methods that concentrate on automatic anomaly detection are not directly capable to assess the pavement condition [5]. The main contribution of this study is to leverage recent advances in deep learning to develop a robust assessment system, capable of detecting, classifying, and characterising pothole automatically. The primary objectives are as follows.

- First, this study introduces a new pothole dataset with depth ground truth values, collected from various roads in Al Khobar city and the KFUPM campus in Saudi Arabia. The dataset has been meticulously annotated manually, ensuring high accuracy and reliability.
- Second, an anomaly segmentation model based on YOLOv8 has been implemented, capable of precisely delineating pothole boundaries across different sizes. This enhances the model's ability to detect road anomalies effectively.
- Finally, this paper provides a comprehensive characterization of detected potholes, including location, area, and depth information. These insights enhance the assessment of pothole severity, aiding contractors and municipal authorities in prioritizing and scheduling road maintenance tasks efficiently.

The remainder of this article is organized as follows. Section 2 reviews related literature on the datasets and techniques employed. Section 3 details the dataset preparation, collection, and monocular depth estimation processes. In Section 4, we describe the proposed methodology for pothole segmentation. Section 5 presents the performance evaluation of model inference and pothole characterization. Finally, Section 6 concludes the paper with a summary of our findings and suggestions for future research.

## 2 Literature Review

Road anomalies pose significant risks, including passenger discomfort, vehicle damage, and even accidents [11]. Ensuring road safety requires timely detection and marking of these anomalies for maintenance. The Internet of Things (IoT) enables seamless interaction between technology and urban infrastructure, allowing roads and cities to communicate and adapt to their environment [12]. One prominent IoT application in smart cities is mobile crowdsensing (MCS), which leverages sensor-equipped devices to collect real-time environmental data [13]. Implementing MCS in road monitoring systems enhances the detection of road anomalies, facilitating efficient maintenance planning. This project builds upon previous research [14] by developing a model capable of estimating pothole depth and analyzing severity based on dimensional characteristics. Existing depth estimation methods have relied on physics-based models, such as Snell's Law for dry and water-filled potholes [15], or photogrammetric techniques [16]. Additionally, some GitHub repositories contain models for pothole depth estimation in India; however, these datasets lack transparency regarding data collection methods, raising concerns about credibility. Moreover, there is currently no open-source, reliable dataset or model for pothole depth estimation.

### 2.1 Existing Datasets:

There were previous studies related to pothole detection and others related to depth estimation. A few publicly available datasets for pothole detection such as RDD2022 [6], which was released to address the Crowd sensing-based Road Damage Detection Challenge (CRDDC 2022) and contained almost more than 47,000 road images where collected from Japan, India, Czech Republic, Norway, the United States, and China, contained almost 55,000 instances of road damages, including cracks and potholes. The RDD2022 dataset was released and annotated for Deep Learning (DL) applications, specifically You Only Look Once (YOLO) algorithm. Additionally, there are datasets that are focused on detecting more general road anomalies. For example, the authors in [7] released a dataset captured in Pakistan including five main types of road anomalies, including: vehicle accidents, vehicle fire, fighting, snatching, and potholes. While this dataset does include pothole classification, it is not the main target of this research, which is identifying accidents and anomalies for security CCTV systems, rather than identifying and classifying potholes accurately. Additionally, there is a public dataset and a model for pothole depth estimation on GitHub gathered by a participant in a Road Safety Hackathon in India, but it only estimates the height, without considering the width or length, and does not explain how the dataset was collected and what are the metrics used to evaluate the model, with no ground truth, which undermines the credibility of using this work. However, there are not any publicly available datasets that provide pothole depth-estimation features that can be predicted using DL. Even though other datasets like the KITTI dataset contained 95,000 highly accurate depth estimation images collected using LiDAR and stereo cameras, it was not intended for estimating the depths or severities of potholes.

Therefore, in this research, we gathered our own dataset using infrared cameras for predicting the depth of potholes after detecting it through a YOLO model and extracting our Region of Interest (RoI). Other research for estimating pothole sizes existed, but most relied on using in-vehicle technologies or advanced cameras for this task. Our task, however, is to be able to infer that from an RGB image input. For example, the authors in **Error! Reference source not found.** developed a YOLOv5 model for predicting potholes, and then utilized existing in-car technologies such as the Lane Keeping Assistance (LKA) system to find the width and length of the detected pothole in millimeters, but it did not detect the depth or the height of the detected pothole. Another related work done by [9] contains 291 images collected from Mumbai City have been used for pothole detection using Mask Region-Based Convolutional Neural Network (Mask R-CNN), and then after extracting the RoI, the area of that pothole is detecting from the distance and pixel size, and achieved an accuracy of 90% between the predicted and the ground truth values with $a \pm 10\%$ deviation. However, these area predictions might not scale well with different camera systems, due to different environments and given the dataset size on which this approach was done. Additionally, it does not provide context regarding the depth or height of the predicted potholes. Furthermore, [10] demonstrated another approach for detecting potholes before the popular use of Computer Vision (CV), which relied mainly on GPS sensors, vibrations, and accelerometers to predict whether potholes exist or not and cluster these data. The study showed great results, especially without using DL approaches, but rather simple Machine Learning (ML) algorithms with a False Positive (FP) rate of 0.2% only. However, the research mainly focused on detecting potholes without visual information such as images, and did not provide information as well regarding its severity. Table 1 summarizes the existing datasets for road anomalies.

Table 1. Summary of previous research papers and datasets on potholes and road anomaly detection.

| Title | Research/Dataset | Details |
|---|---|---|
| RDD2022 Error! Reference source not found. | Research & Dataset | More than 47,000 pictures of potholes with labels for YOLO detection gathered across 7 countries |
| Comprehensive Dataset for Detecting Road Anomalies in Diverse Real-World Situations [7] | Research & Dataset | Images containing general road anomalies such as vehicle accidents, vehicle fire, fighting, snatching, and potholes |
| Depth Estimation of a Pothole on Roads on GitHub | Dataset | Images containing potholes with depth estimation as labels. However, there is no information regarding the methodology or the validity of the dataset. |
| KITTI Dataset | Dataset | A dataset of 95,000 images collected using multiple sensors. LiDAR and stereo cameras were used for depth estimation images, but the dataset was not dedicated for pothole detection and dimensions estimation. |
| Augmenting roadway safety with machine learning and deep learning: Pothole detection and dimension estimation using in-vehicle technologies [8] | Research | Uses in-built vehicle technologies and systems such as LKA to detect pothole and then approximate its dimensions in width and length, but not in height. |
| Deep Learning Model for Pothole Detection and Area Computation | Research | 291 images of potholes collected in Mumbai City, and then the area is approximated. However, no approach for predicting the depth is covered. |
| The pothole patrol: using a mobile sensor network for road surface monitoring [10] | Research | Pothole detection using GPS sensors, vibrations, accelerometers, and then simple ML algorithms and clustering were used for prediction. However, it was only focused on pothole detections with no visual information, but it was ahead of its time. |

To address this gap, this project manually collects a high-quality dataset using Microsoft Kinect V2, a stereo camera equipped with an infrared sensor, projector, and RGB camera. The Kinect camera generates heatmaps, which provide depth measurements of road anomalies. These ground-truth depth values will serve as training data for a semantic segmentation model, enabling precise pixel-wise depth estimation of potholes and road irregularities. By leveraging this approach, the project aims to enhance road safety through accurate anomaly detection and severity assessment.

# 3 Data Collection and Preparation

This paper focuses on automatic identification and characterization of pothole anomaly in roads using digital images captured from a dashboard-mounted camera. A new dataset of images and their corresponding depth maps was captured from diverse roads from Al Khobar city and KFUPM campus in Saudi Arabia. The images were firstly augmented to ensure that the model can generalize better on unseen data, making it robust for real-world applications, and then manually annotated. The quality of these annotations directly impacts the model's accuracy and performance, as improper labelling could lead to erroneous predictions. It consists of 981 images with size of 1920x1080 pixel. They were resized to 640x640 pixels to standardize the input for the model. The dataset was divided into 931 images for training and validation and 50 images for testing.

The dataset is composed of pothole images. Figure 1 shows the data collection setup. The ground truth was collected using a Microsoft Kinect camera **Error! Reference source not found.** mounted on a tripod, a trolley, and a laptop. A DC power supply was connected to an AC converter which was used to power the setup. To replicate the actual implementation, the tripod was used at two different heights. One height is for the average distance from the ground to the hood of a sedan car, 83 centimetres, and another image is taken to replicate the environment of an SUV, at a height of 112 centimetres. Multiple photos were taken at different angles and different distances to prevent model overfitting. Figure 2 shows a sample pothole as captured by the setup (4 sides + ground truth).

## 3.1 Dataset Description

During our experiments, we used our own dataset. We collected 247 RGB images and their depth maps from multiple streets.

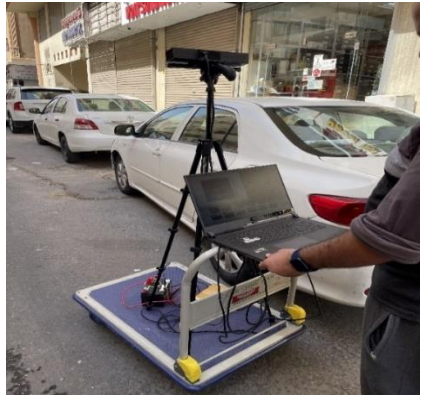

Figure 1: the setup used for dataset collection.

In this stage, the data is first cleaned by deleting all images that consist of only black pixels, all-zero images. This might be caused due to the fact that the program needs a few seconds to set up the Kinect, and sometimes the Kinect is not fully set up after the grace period. After that, the images in the dataset are augmented to reduce overfitting. Each image will have at least four extra augmented images:

- Saturation changed image
- Inverted image – the depth scan is inverted as well
- Saturation changed and inverted image – the depth scan is inverted as well
- Random augmentations including brightness, contrast, rotation, flip, and saturation

Applying these transforms resulted in a total of 981 image pairs, each consisting of an RGB image and its corresponding depth map. Out of these, 931 pairs were allocated for training and 50 pairs for testing.

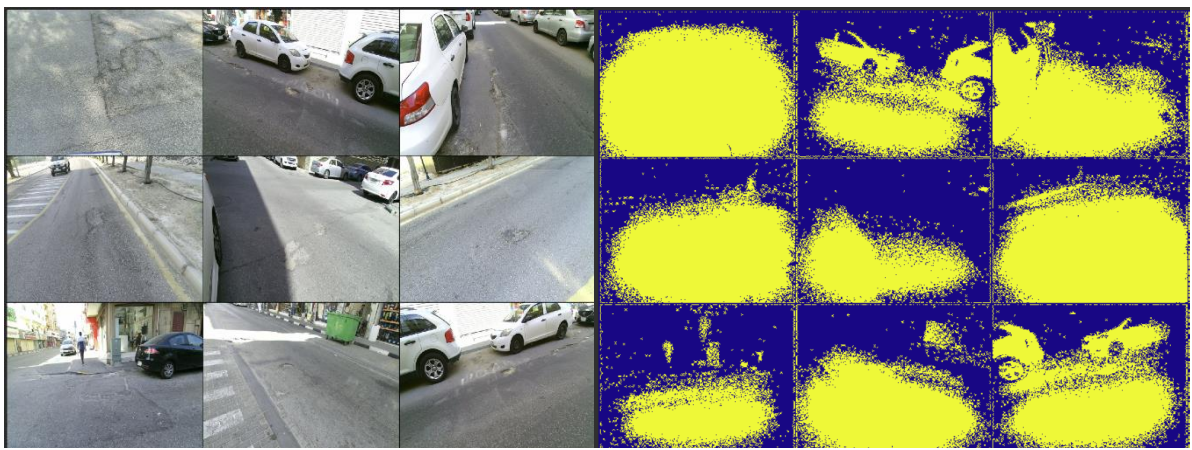

Figure 2: Samples of our collected dataset, RGB and depth pairs.

As can be seen in Figure 2, there are some depth pixels values missing. Since we already have the depths of the anomalies (our intended targets), we did not interpolate the missing pixels in the collected depth maps.

## 3.2 Training Models

We used a workstation with an AMD Ryzen Threadripper, Nvidia RTX A6000, and a Python 3 environment. We trained two Encoder-Decoder models on the collected RGB-Depth pairs.

### 3.2.1 DenseDepth Model [6]:

Since our dataset is small, we trained the DenseDepth model for one thousand epochs with a batch size of 16. We used an initial learning rate of 0.0001 and it decreases after each epoch. Figure 3 and Figure 4 show the training and validation loss for the DenseDepth model. Figure 5 shows the predicted depths while training and the difference between predicted and actual depths for the DenseDepth model.

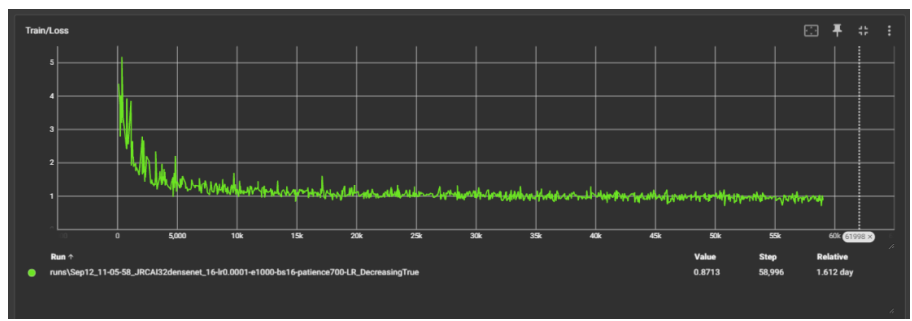

Figure 3: Training loss for the DenseDepth model.

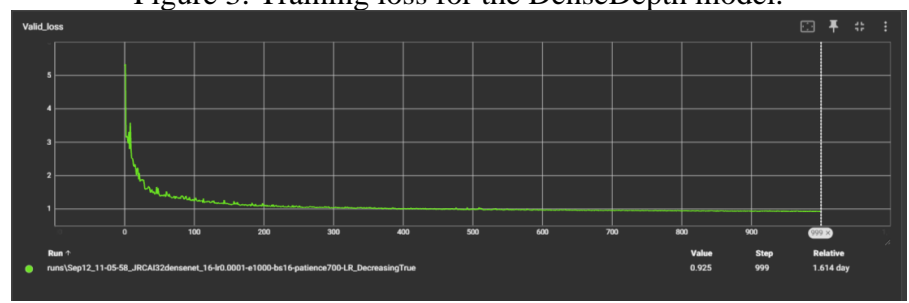

Figure 4: Validation loss for the DenseDepth model.

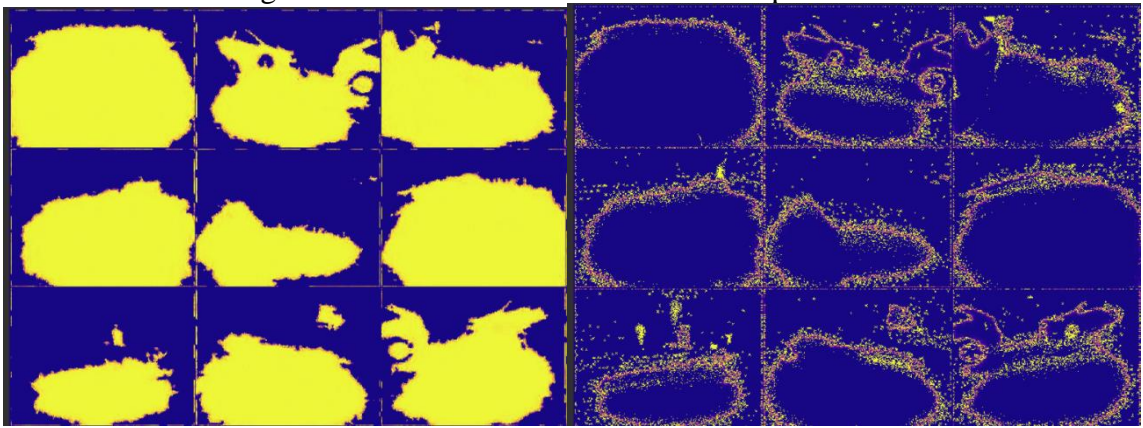

Figure 5: Predicted depth (on the left) and the difference between predicted and actual depths (on the right) for the DenseDepth model.

1- Our Model **Error! Reference source not found.**:

We trained the encoder decoder model for one thousand epochs with a batch size of 16. We used an initial learning rate of 0.0001 and it decreases after each epoch. Figure 6 and Figure 7 show the training and validation performance metrics for our model.

Figure 6: Training performance metrics for our model.

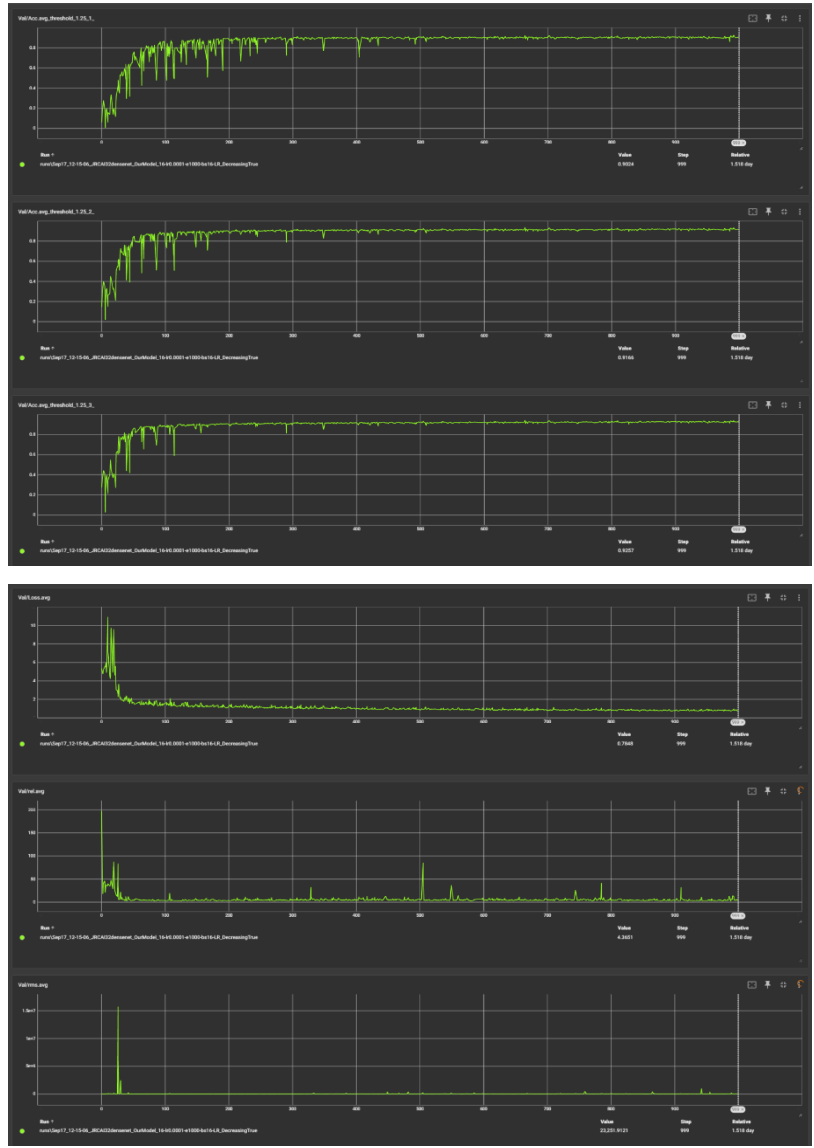

Figure 7: Validation performance metrics for our model.

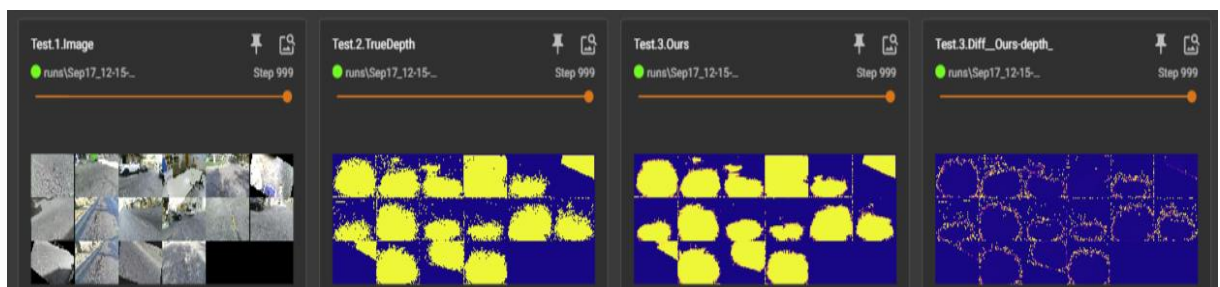

Figure 8: The inputted RGB images (first), the ground truth depth maps (second), the predicted depth (third), and the difference between predicted and actual depths (fourth) for our model.

### 3.2.2 Testing

We tested the trained models on 50 unseen RGB and depth pairs. The average RMSE for the predicted depths using the DenseDepth model is 1.25, while by our model is 1.74. It is most likely due to selecting the testing samples randomly, or our second depth estimation model is not differentiating easily between close depths (very small differences in depths).

# 4 Transfer Learning for Pothole Segmentation

Transfer learning was employed to enable rapid adaptation to the unique characteristics of the pothole dataset without requiring training from scratch. The YOLO family of models is favored for its real-time object detection capabilities and ease of training compared to other deep learning models [19]. In particular, YOLOv8-seg offers detailed segmentation while maintaining fast inference speeds, making it ideally suited for real-world road damage assessments [20].

This work proposes the use of YOLOv8-seg for pothole detection and segmentation analysis. The objective is not only to identify potholes but also to segment them accurately, thereby providing valuable spatial information about the extent of the damage and supporting enhanced road maintenance initiatives. The model's ability to be fine-tuned on specific road damage data allows for high performance with minimal computational resources, which is crucial for deployment on edge devices.

Among the five pre-trained YOLOv8-seg models, YOLOv8n-seg—the smallest and fastest—strikes a balance between speed and accuracy, making it particularly suitable for real-time applications. Given the constraints of edge device deployment, this research focuses on smaller, faster models such as YOLOv8n-seg and YOLOv8s-seg, as detailed in Table 2 [19].

**Table 2:** Comparison of different scale-seg models

| Model | Depth | Width | Parameters |
|---|---|---|---|
| Yolov8n-seg | 0.33 | 0.25 | 3.26 |
| Yolov8s-seg | 0.33 | 0.50 | 11.79 |
| Yolov8m-seg | 0.67 | 0.75 | 25.89 |
| Yolov8l-seg | 1.00 | 1.00 | 42.90 |
| Yolov8l-seg | 1.00 | 0.25 | 67.0 |

## 4.1 Model Performance Analysis (Evaluation Metrics)

This section evaluates the effectiveness of the YOLOv8n-Seg model through widely used statistical measures such as precision, recall, F1 score, and mAP. The mathematical equations of precision (1), recall (2), F1 score (3), and mAP (4) are described as bellow.

$$Precision = \frac{T_P}{T_P + F_P} \quad (1)$$

$$Recall = \frac{T_P}{T_P + F_N} \quad (2)$$

$$F1 = \frac{2 * Precision * Recall}{Precision + Recall} \quad (3)$$

$$mAP = \sum_{n=1}^{K} \frac{AP_i}{K} \quad (4)$$

where, $T_P$ is True Positive (a pothole identified as a pothole), $T_N$ is True Negative (a non-pothole identified as a non-pothole), $F_P$ is False Positive, (a non-pothole identified as a pothole), and $F_N$ is False Negative, (a pothole identified as a non-pothole).

# 5 Results

This section presents the stages of the conducted ablation study to optimize the performance of road anomaly characterisation system from static RGB image data. The section discusses the outcomes of training different YOLOv8 models using new collected datasets.

**Table 3:** Initialized parameters of Yolov8 model.

|   | Parameter | Value |
|---|---|---|
| 1 | optimizer | AdamW |
| 2 | Initial learning rate | 0.002 |
| 3 | Number of epochs | 150 |
| 4 | patience | 50 |
| 5 | Batch | 16 |
|   | Momentum | 0.9 |
| 6 | Image size | 640*640 |
| 7 | Decay |  |

Figure 9 displays loss patterns in training and validation over epochs, which demonstrates the model's learning progress during the training and validation processes. The losses also indicate the accuracy of the model in predicting the boxes and segment the images throughout the training and validation phases. While the classification losses (cls) illustrate the model's capability to classify the objects correctly within the bounding boxes. The distribution focal loss (dfl) indicates how well the model is learning to recognize, classify, and segments potholes in the images, particularly focusing on the challenging cases.

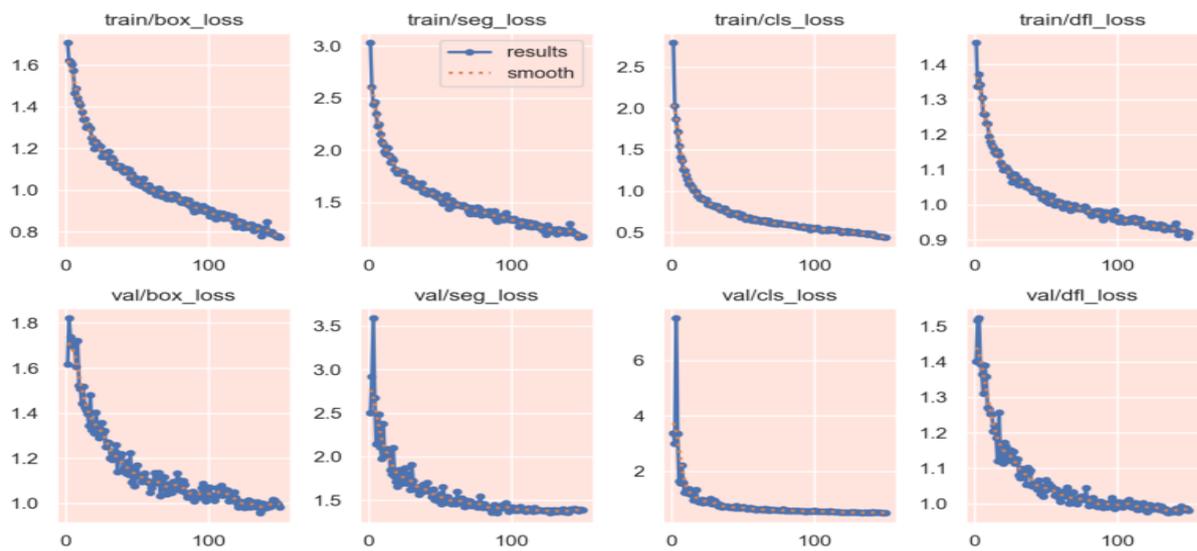

Figure 9: Training and Validation Loss Trends (150 epochs).

Figure 10 presents the precision metrics for both bounding box (B) and mask (M) predictions, which indicate the proportion of correct positive predictions made by the proposed model. Meanwhile, the recall metrics assess the model's ability to identify all relevant instances within the dataset. The mean average precision (mAP) values for both bounding box and mask predictions summarize the accuracy at specific Intersection over Union (IOU) thresholds (IOU=0.50 and IOU=0.50-0.95).

Figure 11 illustrates the training progress, showing that the training and validation losses are closely aligned—a positive sign that the model generalizes well to unseen data. Although the slight difference between training and validation losses suggests a minor degree of overfitting, the model still performs well in predicting bounding boxes. Additionally, the strong classification performance observed on both the training and validation datasets indicates that further improvements in this aspect are unlikely.

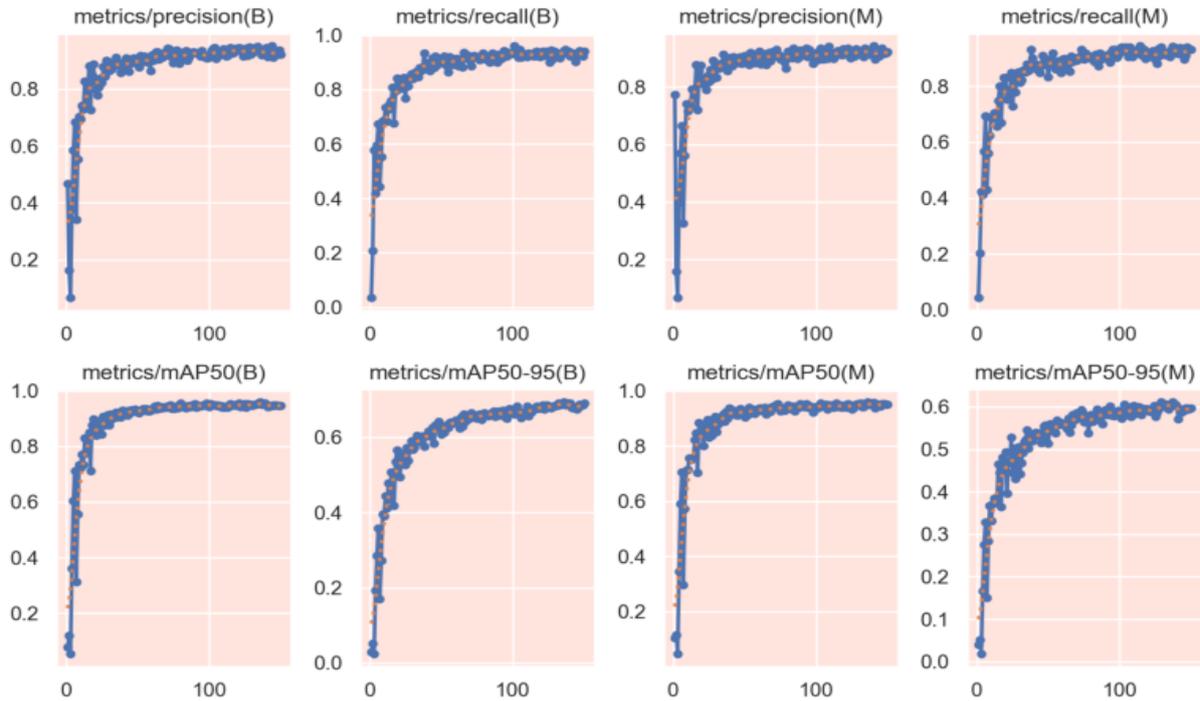

Figure 10: Training and Validation Loss Trends (150 epochs).

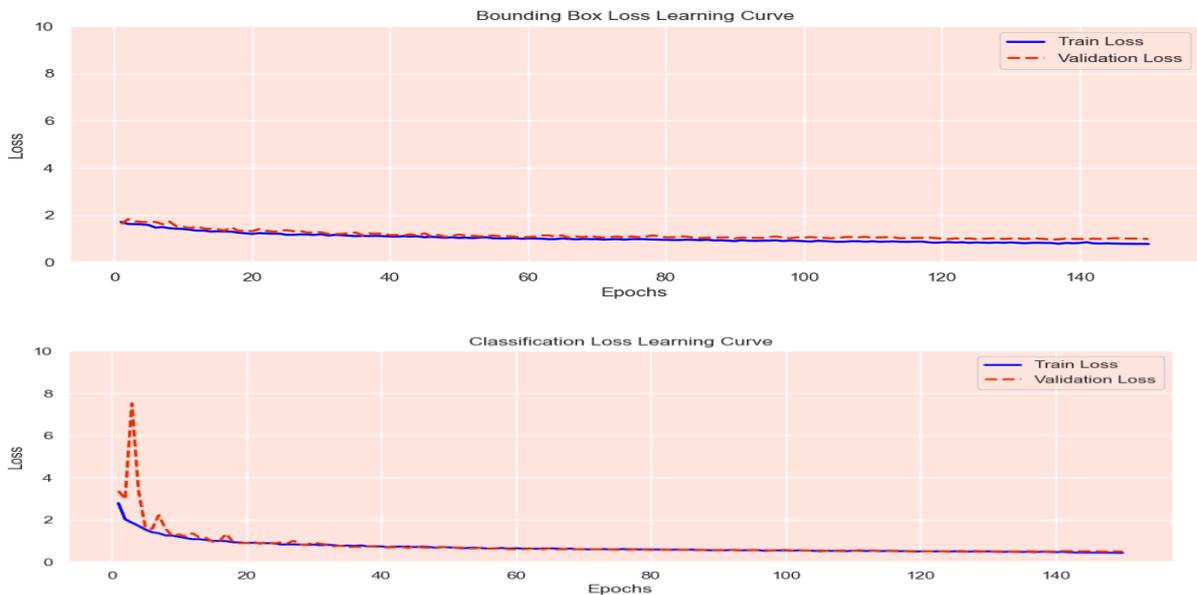

Figure 11: Bounding box and classification loss learning curves (150 epochs).

In Figure 12, the Distribution Focal Loss demonstrates that the model focuses on challenging instances, with the validation performance suggesting that a bit of additional tuning could be beneficial. The segmentation loss further confirms the model's strong capability in segmentation, as evidenced by the minimal variability in validation loss. The close alignment of these loss curves indicates good generalization. Incorporating a more diverse training dataset may further enhance the model's overall performance.

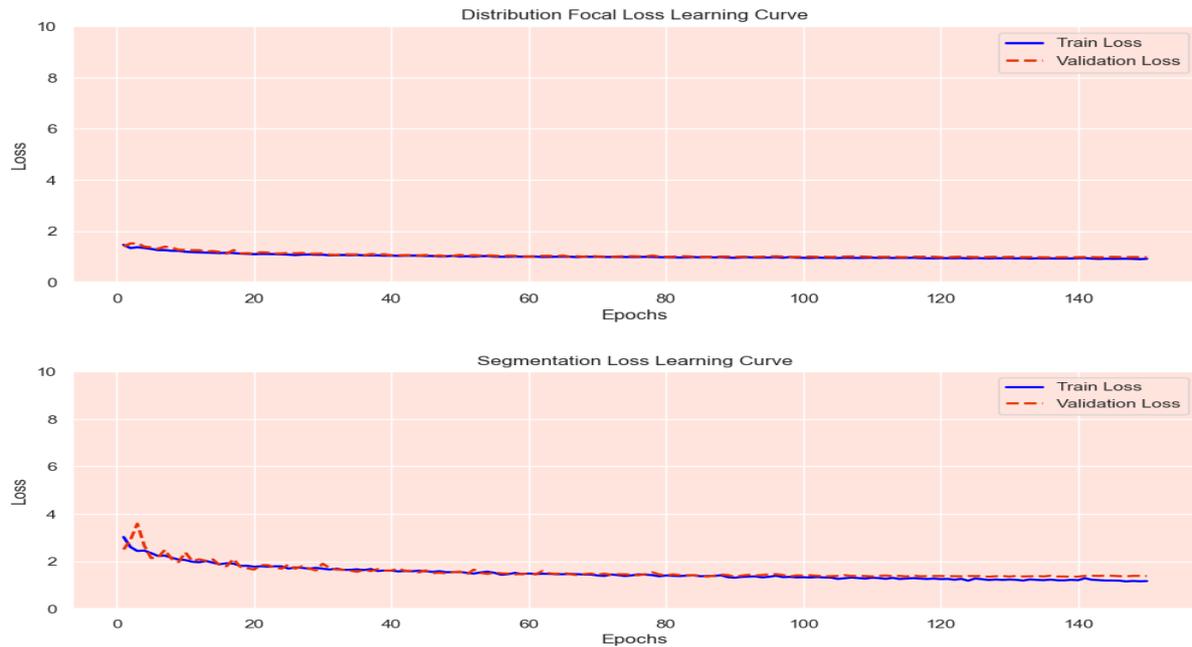

Figure 12: Distribution focal and segmentation loss learning curves (150 epochs).

Figure 13 illustrates precision, recall, and F1 score confidence curves that reflect the exceptional performance of the proposed model. The model maintains near-perfect precision across all confidence levels for both bounding box and mask predictions, demonstrating its accuracy. The precision-recall curves reveal that, despite the inherent trade-off, the model sustains a high recall across a range of confidence thresholds, consistently capturing true positives. Additionally, the stable and high F1 scores highlight the balance between precision and recall for both prediction types, even at higher confidence levels. Overall, these consistently high metrics indicate that the model generalizes well and exhibits robust predictive performance, requiring minimal adjustments for practical deployment.

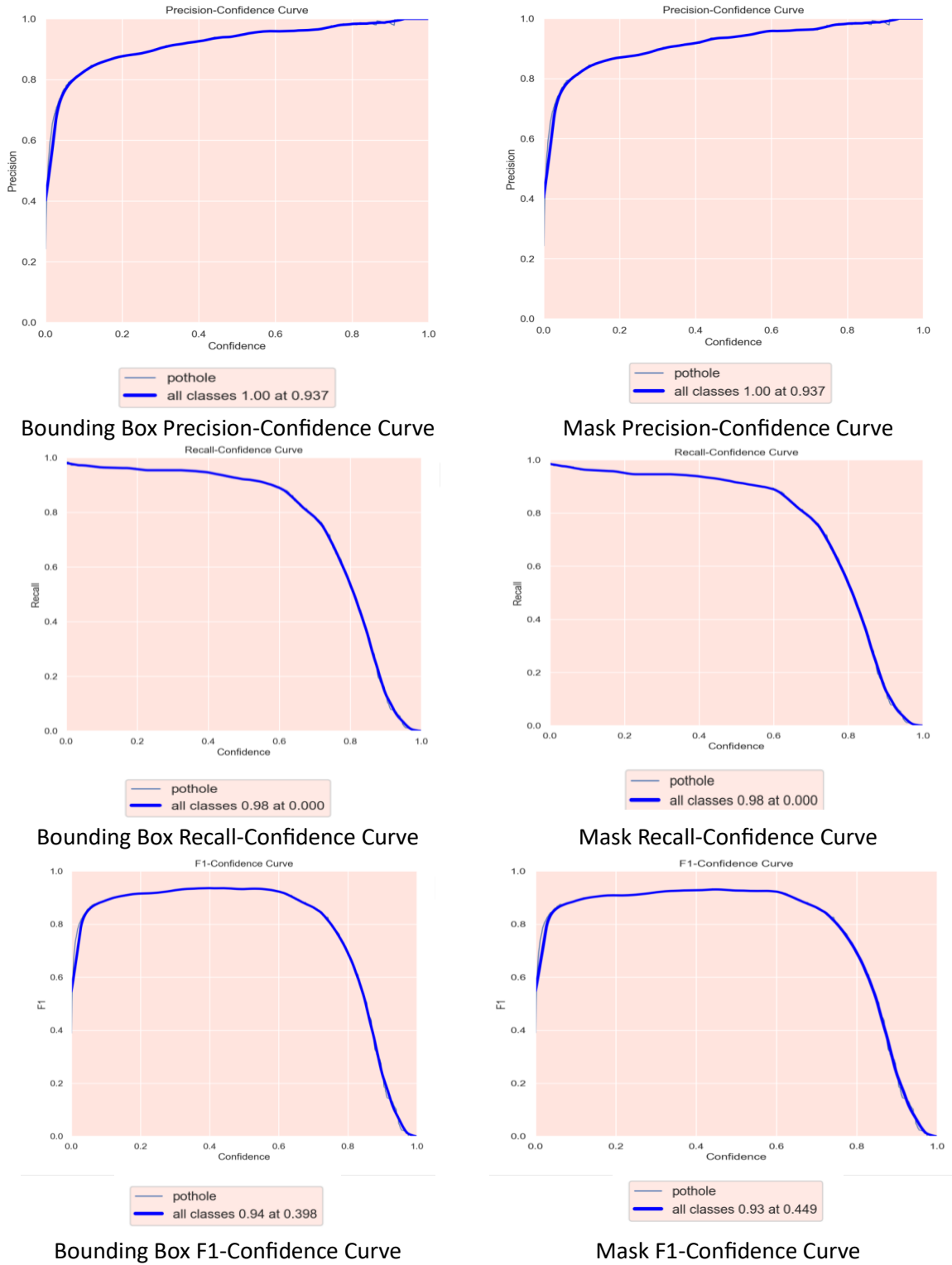

Bounding Box Precision-Confidence Curve    Mask Precision-Confidence Curve

Bounding Box Recall-Confidence Curve    Mask Recall-Confidence Curve

Bounding Box F1-Confidence Curve    Mask F1-Confidence Curve

Figure 13: Precision, recall, and F1 confidence curves of bounding box and mask.

The precision-recall curves are beneficial for observing the trade-off between precision and recall for various confidence thresholds. Figure 14 indicates a high level of precision, evidenced by a mean Average Precision (mAP) of 0.946 and 0.953, in bounding box and mask of potholes, indicating the model's effectiveness in accurately identifying and segmenting roads' potholes. The high precision across various recall levels indicates the model reliability for practical use.

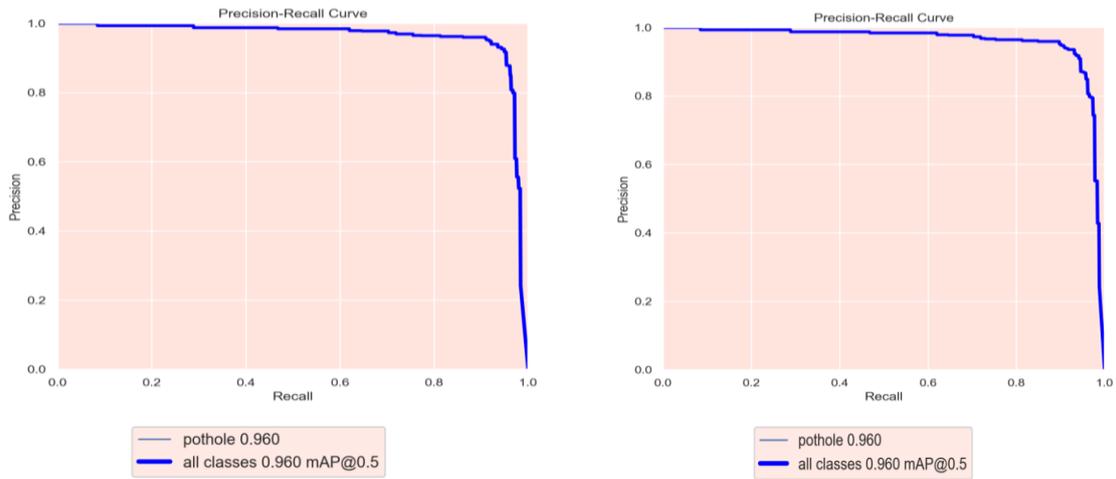

Bounding Box Precision-Recall Curve                    Mask Precision-Recall Cure

Figure 14: Precision-recall curves

The confusion matrix in Figure 15 displays the true potholes and background against the predicted ones.

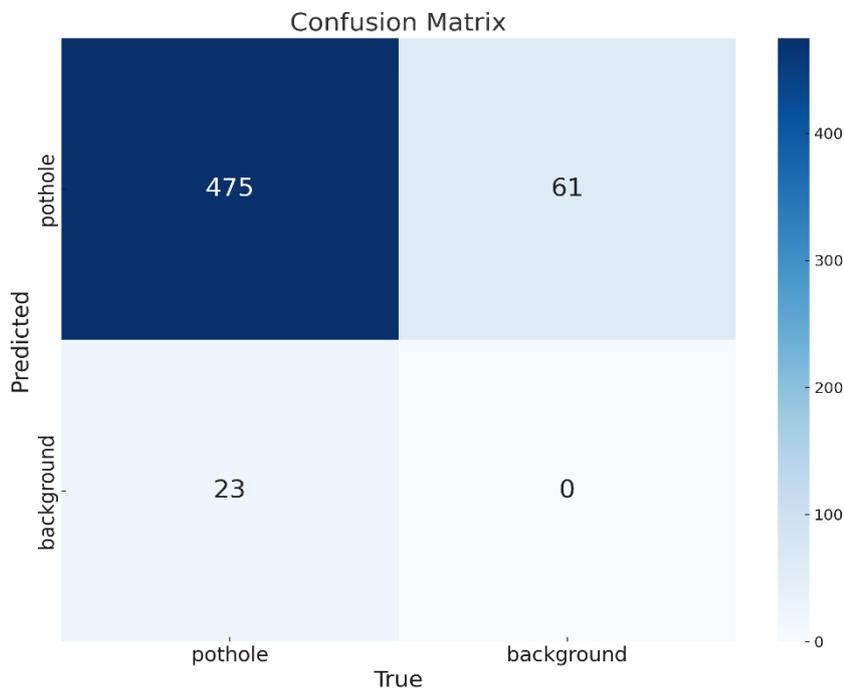

Figure 15: Confusion Matrix.

Table 3: Overall validation performance Metric Assessment.

|  | Box | | | | Mask | | | |
| --- | --- | --- | --- | --- | --- | --- | --- | --- |
| Models | P | R | mAP50 | mAP50-95 | P | R | mAP50 | mAP50-95 |
| Yolov8n | 0.93 | 0.944 | 0.96 | 0.694 | 0.925 | 0.932 | 0.96 | 0.606 |

## 5.1 Model Inference and Pothole Characterisation

The performance of the developed model was initially evaluated on a subset of images from the validation set. It was then tested on an independent sample of images to assess its accuracy in detecting and segmenting potholes. Finally, the area of each detected pothole was calculated by analyzing the contours on the segmentation mask, allowing for the determination of the total damaged area and the percentage of road damage attributable to potholes. Figure 16 highlights the capabilities of the best-performing segmentation model using selected images from the validation set.

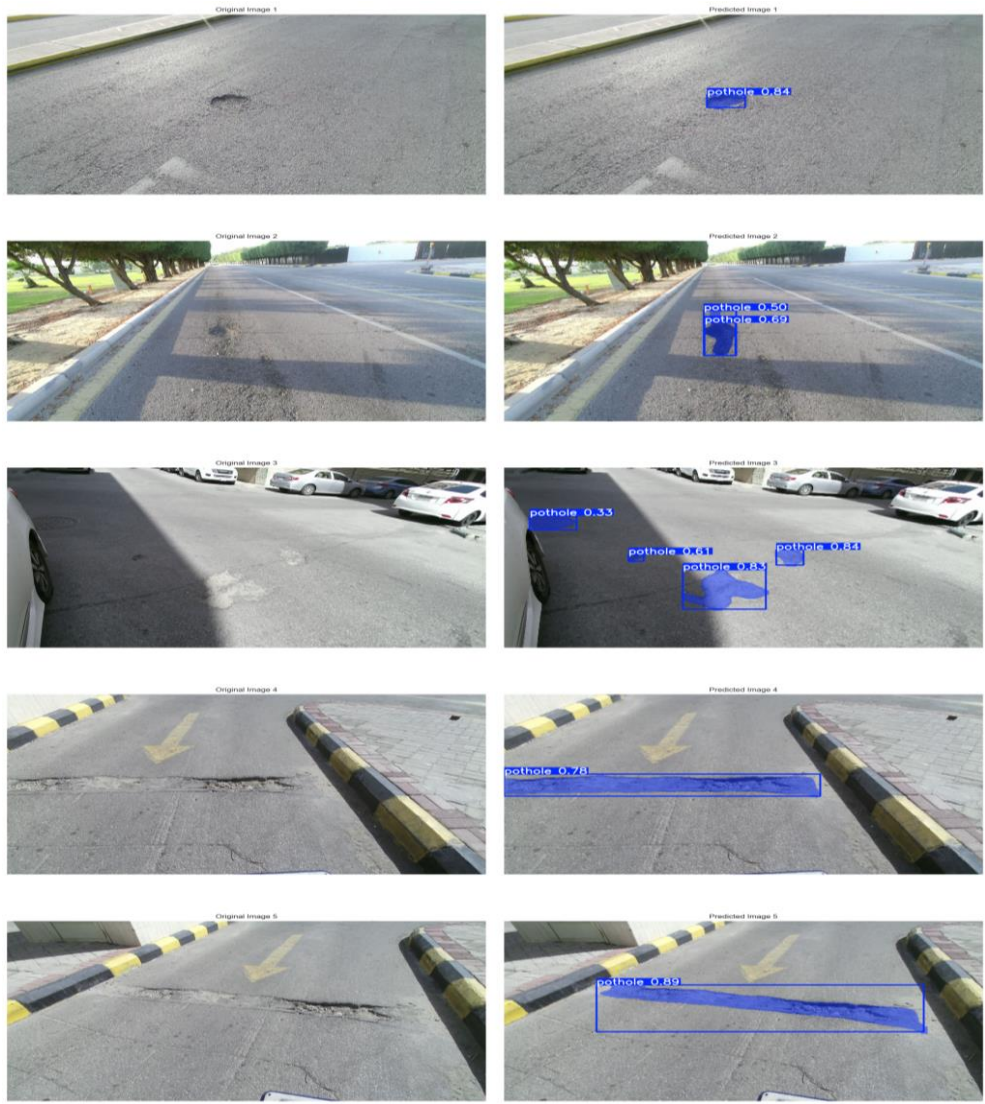

Figure 16: Test set inferences.

The results from the segmentation and depth estimation models were combined and evaluated on a comprehensive dataset to assess their effectiveness in characterizing road potholes—an effort aimed at enhancing road maintenance and safety. This characterization process involved detecting potholes, accurately pinpointing their locations, calculating their area and depth, and quantifying the percentage of road damage caused by pothole-affected regions. Several illustrative examples are presented below.

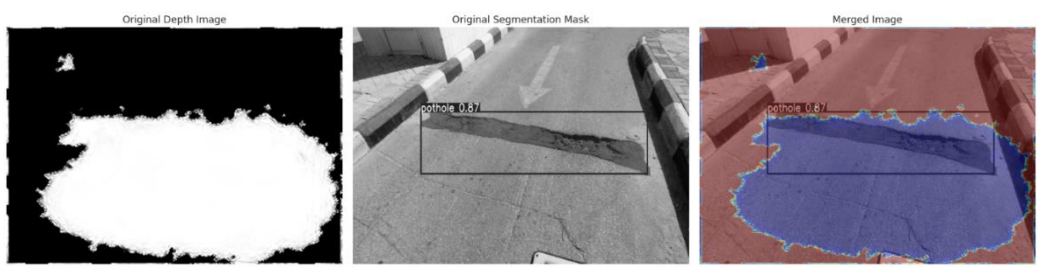

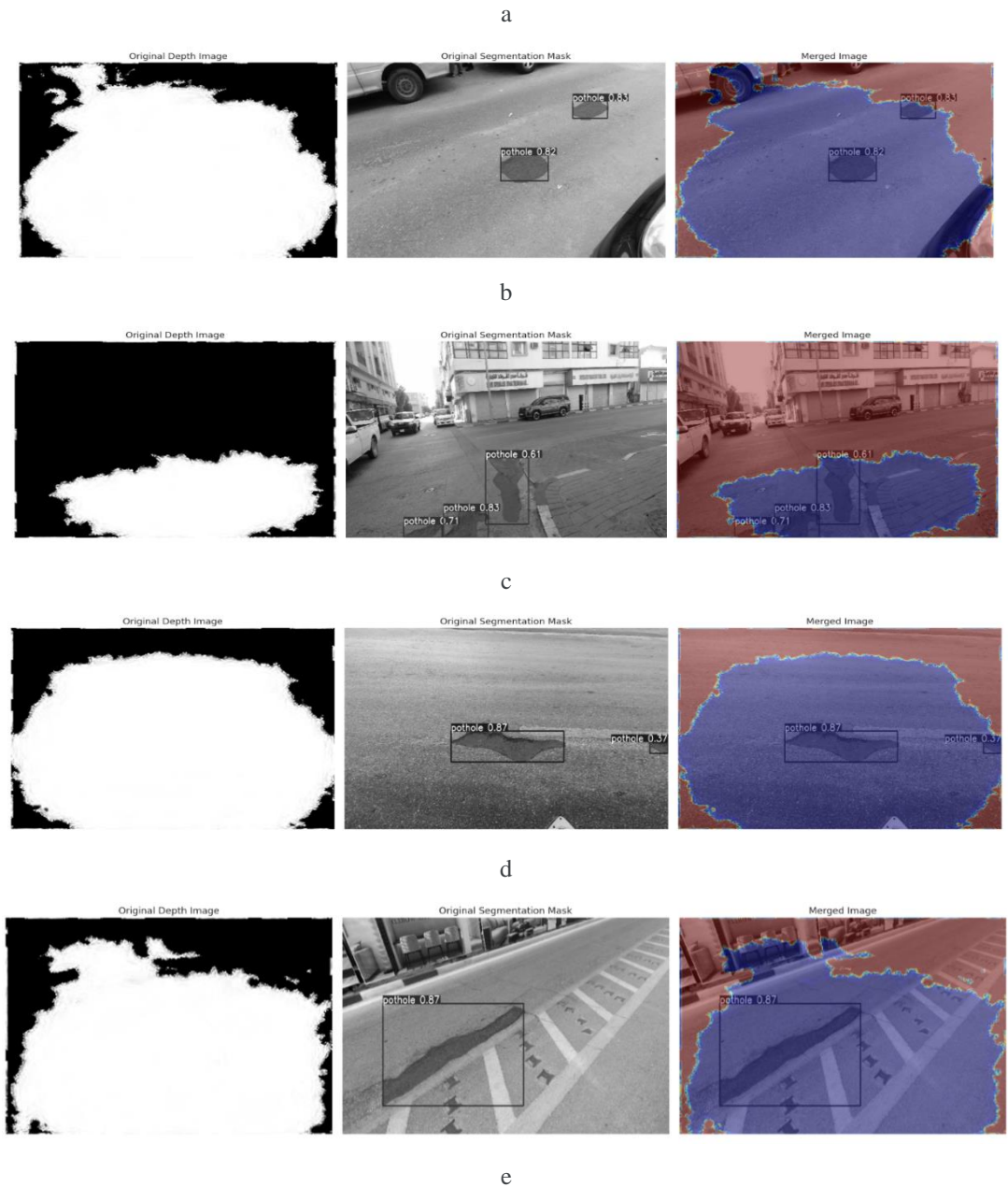

Figure 17: Depth map and segmented images, and their merged image a, b, c, d, e.

| Number of Example | Number of potholes | Pothole area (pixels) | Total area (pixels) | Percentage of damage (%) |
|---|---|---|---|---|
| a | 1 | 12101.0 | 245760 | 4.92 |
| b | 2 | 1723.5 + 3778.0 = 5501.5 | 245760 | 2.24 |
| c | 3 | 2497.0 + 1113.5 + 43.0 = 3653.5 | 245760 | 1.49 |
| d | 1 | 5929.0 | 245760 | 2.41 |
| e | 1 | 8923.0 | 245760 | 3.63 |

In this section, we go further in characterizing the anomaly (i.e., potholes) by estimating the relative pothole depth ($RP_D$), which is defined as the ratio of the average depth of surrounding area to the average depth of pothole area. This approach is restricted to include only the surrounding pixels and not all the surrounding area to ensure more accurate analysis of the pothole depth.

Figure 18a shows that the depth of pothole area ($P_D$) is 0.7693, while the average depth of surrounding area ($S_D$) is 0.5808. we can observe the following:

  a- $P_D > S_D$, the depth estimation algorithm succeeds in detecting the pothole depth!
  b- $RP_D = \frac{P_D - S_D}{S_D} * 100 = 18.85\%$ ; this measure along with the pothole area provides an indication of the danger of the detected pothole.

Following the same procedure in Figure 18b, we find that that depth of pothole area ($P_D$) is 0.6925, while the average depth of surrounding area ($S_D$) is 0.5254. we can observe the following:

  a- $P_D > S_D$, the depth estimation algorithm succeeds in detecting the pothole depth!
  b- $RP_D = \frac{P_D - S_D}{S_D} * 100 = 16.71\%$ .

Applying the same procedure, we notice that the $RP_D$ in Figure 18c is equal to 8.46%; this measure along with the pothole area indicates that this pothole less dangerous than the ones in Figure 18(a, b).

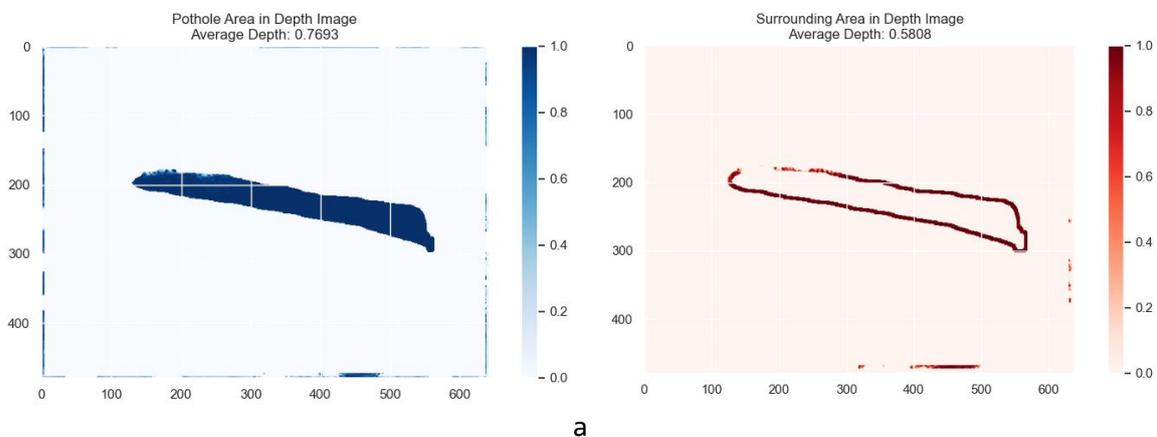

a

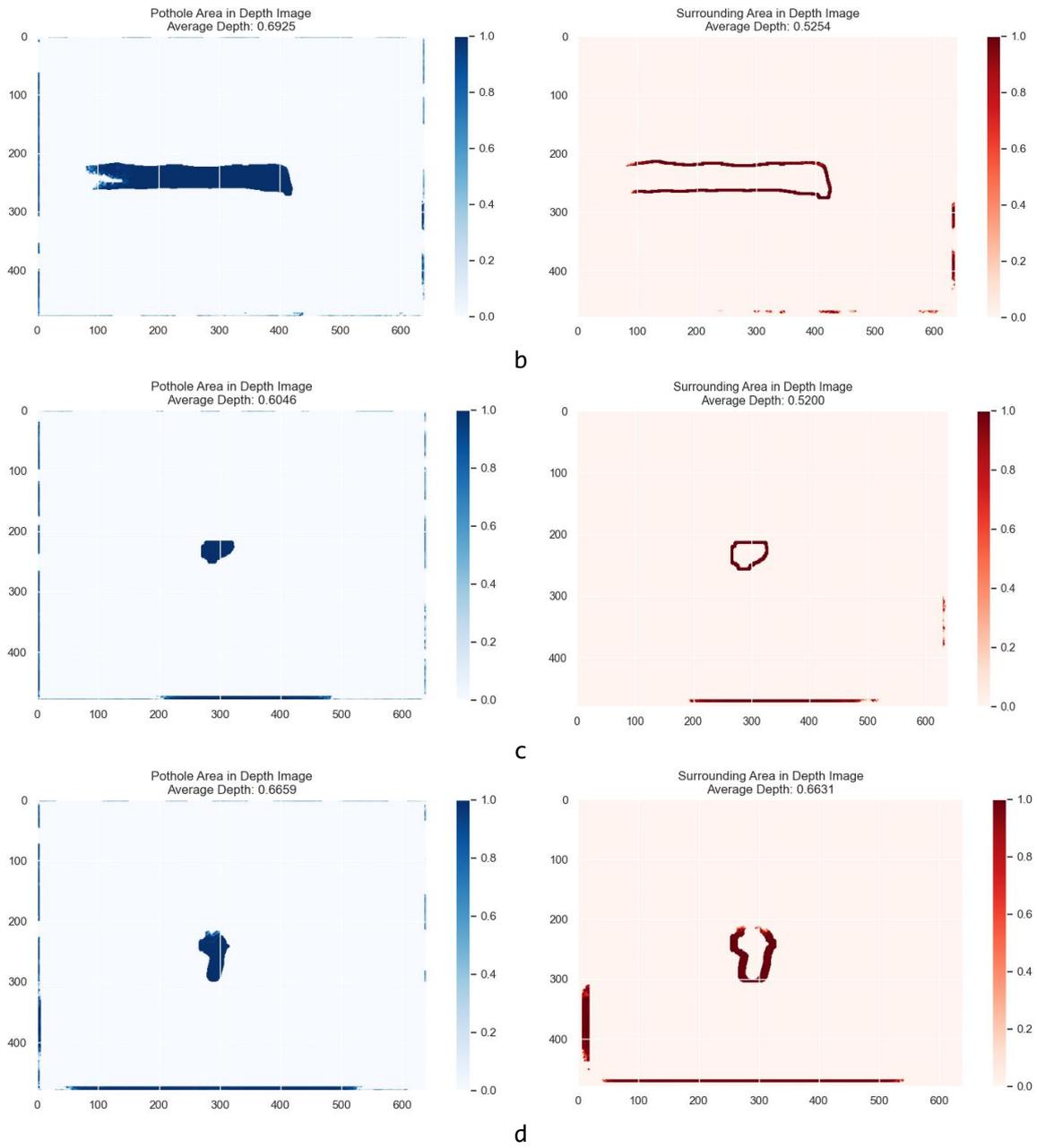

Figure 18: Samples of detected potholes and their average depth.

| Figure 18 | Average Depth (normalized units) | |
|---|---|---|
| | Pothole Area | Surrounding Area |
| a | 0.7693 | 0.5808 |
| b | 0.6925 | 0.5254 |
| c | 0.6046 | 0.5200 |
| d | 0.6659 | 0.6631 |

## 6 Conclusion

Deployment of machine learning for automated pavement distress detection is no longer a novel concept; however, the application of deep learning continues to captivate pavement researchers and practitioners. While several publicly available datasets for road anomalies exist, most focus solely on the external appearance of the anomalies and lack the detailed depth information required for comprehensive characterization. To address this gap, our work began by developing and collecting a new dataset specifically designed to capture the depth of potholes. This dataset comprises 1000 RGB images along with their corresponding ground truth maps.

Leveraging this dataset, we trained two depth models to generate accurate depth maps for unseen potholes, and employed the YOLOv8-seg model for robust pothole detection and segmentation. This integrated approach not only enhances road safety and maintenance practices by pinpointing pothole locations but also provides a comprehensive characterization by calculating both the area and depth of each pothole.

With ongoing advancements in computer vision and deep learning, pothole detection systems hold significant potential to minimize accidents and improve overall road safety. By enabling authorities to precisely identify areas in need of repair and alerting drivers to hazardous road conditions, these systems can play a crucial role in efficient road maintenance and planning, ultimately ensuring smoother and safer travel.